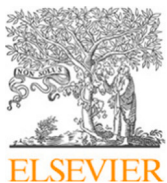
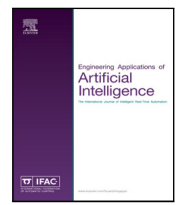

Research paper

# Gait-based age group classification with adaptive Graph Neural Network

Timilehin B. Aderinola [c], Tee Connie [a,*], Thian Song Ong [a], Andrew Beng Jin Teoh [b], Michael Kah Ong Goh [a]

[a] *Faculty of Information Science and Technology, Multimedia University, Malacca, Malaysia*
[b] *School of Electrical and Electronic Engineering, College of Engineering, Yonsei University, Seoul, South Korea*
[c] *School of Computer Science, University College Dublin, Ireland*

A B S T R A C T

Deep learning techniques have recently been utilized for model-free age-associated gait feature extraction. However, acquiring model-free gait demands accurate pre-processing such as background subtraction, which is non-trivial in unconstrained environments. On the other hand, model-based gait can be obtained without background subtraction and is less affected by covariates. For model-based gait-based age group classification problems, present works rely solely on handcrafted features, where feature extraction is tedious and requires domain expertise. This paper proposes a deep learning approach to extract age-associated features from model-based gait for age group classification. Specifically, we first develop an unconstrained gait dataset called Multimedia University Gait Age and Gender dataset (MMU GAG). Next, the body joint coordinates are determined via pose estimation algorithms and represented as compact gait graphs via a novel part aggregation scheme. Then, a **P**art-**A**dapt**I**ve **R**esidual **G**raph **C**onvolutional **N**eural Network (PairGCN) is designed for age-associated feature learning. Experiments suggest that PairGCN features are far more informative than handcrafted features, yielding up to 99% accuracy for classifying subjects as a child, adult, or senior in the MMU GAG dataset. These results suggest the feasibility of deploying Artificial Intelligence-enabled solutions for access control, surveillance, and law enforcement in unconstrained environments.

## 1. Introduction

Although the historical origin of age measurement is uncertain, it became precise with the standardization of time and date. Humans can roughly estimate age from physical appearance when it is impossible to obtain an individual's chronological age. However, relying on eyeballing techniques for age prediction in mission-critical applications is not feasible. This motivates research in machine-based age estimation or age group classification based on physical attributes. Application areas include but are not limited to access control, surveillance, law enforcement, and human–computer interaction (Islam et al., 2021).

The two main steps in age group classification are age-associated feature extraction and classifier design. The main challenge is extracting robust age-associated features from physical attributes such as face or gait (Liao et al., 2018). In particular, although it is possible to obtain gait from a distance with video acquisition devices, gait features are affected by covariates such as viewpoint, apparel, walking speed, walking slope, and footwear (Rani and Kumar, 2023). In unconstrained environments, gait can only be acquired using model-based or model-free vision-based techniques. Model-based approaches measure body part parameters such as lengths and range of motion via skeleton representations. On the other hand, model-free approaches extract features from silhouette images obtained after background subtraction. Though more computationally expensive, the model-based approaches are more robust to external covariates such as clothing and carried objects, thanks to advanced pose estimation techniques (Aderinola et al., 2021b).

Deep learning has been widely used for age-associated feature extraction based on model-free gait descriptors such as gait energy images (Berksan, 2019; Lau and Chan, 2023; Russel and Selvaraj, 2021; Xu et al., 2021b; Kitchat et al., 2022; Xu et al., 2021a; Zhang et al., 2018). However, acquiring model-free gait requires accurate background subtraction, which is non-trivial in unconstrained environments. On the other hand, model-based gait can be obtained without background subtraction and is less affected by covariates. However, most model-based gait approaches for age estimation and age group classification (Aderinola et al., 2021a; Begg et al., 2005; Yang and Wang, 2015; Hediyeh et al., 2013; Hema and Pitta, 2019; Punyani et al., 2018; Yoo and Kwon, 2017; Zhang et al., 2010) merely utilize handcrafted features, which requires domain expertise.

**Table 1**
MMU GAG datasets in comparison with similar gait datasets.

| Dataset | # | Age Groups[a] | Scene/Time | Covariates |
|---|---|---|---|---|
| SOTON (Shutler et al., 2004) | 115 | – | In, out/Day | – |
| CASIA-B (Yu et al., 2006) | 124 | *adult* | In/Day | CO, DR. |
| USF 1 (Phillips et al., 2002) | 74 | *adult* | Out/Day | V, TR, SH, CO, T. |
| USF 2 (Sarkar et al., 2005) | 85 | *adult* | Out/Day | V, TR, SH, CO, T. |
| Outdoor-Gait (Song et al., 2019) | 138 | – | Various/Day | CL, scene, CO. |
| GITW (ours) | 439 | *child, adult, senior* | Arbitrary | Naturally diverse. |

#: Number of subjects; In: Indoor; Out: Outdoor; V: view; CO: Carried Object; DR: dressing; SH: shoe; T: time; TR: terrain/walking surface.

[a] *child* = [1–17]; *adult* = [18–64]; *senior* = [64–] years.

Recent emerging studies on Graph Neural Networks (GNNs) generalize Deep Neural Networks (DNNs) to data that cannot be represented in Euclidean space (Bronstein et al., 2017; Zhao et al., 2020; Teepe et al., 2021; Sheng and Li, 2021). Furthermore, with recent advances in human pose estimation techniques, it is now possible to obtain joint coordinates automatically and accurately. Hence, gait can be modeled with joint coordinates as a graph, such as joints as nodes and bones as edges. This allows representation learning on model-based gait with GNNs.

This paper proposes a dedicated Graph Convolutional Network (GCN), Part-AdaptIve Residual Graph Convolutional Neural Network (PairGCN), to extract age-associated features from model-based gait in unconstrained environments, which has not been attempted before. The PairGCN can automatically adjust its edge weights based on the dominant body parts in the input part graph. More specifically, PairGCN learns which body parts are essential in classifying subjects as a child, adult, or senior by taking advantage of our part aggregation scheme.

We also introduce the gait in the wild (GITW) as the first dataset in the Multimedia University Gait Age and Gender datasets (MMU GAG).[1] GITW is naturally diverse with respect to age, race, gender, walking surface, shoe, and dressing. We impose no restrictions based on the scene, age, gender, or ethnicity, except that we restrict the collection to video recordings of healthy individuals walking at a normal pace. Table 1 compares GITW with similar existing gait datasets collected in unconstrained environments. Accordingly, GITW contains more subjects, covers more age groups, and is more naturally diverse in terms of dressing, walking surface, gender, and recording time.

Our main contributions are itemized below.

1. We study age group classification problems using model-based gait features determined by the deep learning approach, which is the first of its kind.
2. A **P**art-**A**dapt**I**ve **R**esidual GCN (PairGCN) is proposed for automatic extraction of age-associated gait features in video sequences, which can incorporate age-associated part information from model-based gait.
3. A gait age and gender dataset, Gait in the Wild (GITW) is collected, pre-processed, analyzed, and released for age group classification in unconstrained environments.

The overall structure of this study is shown in Fig. 1. The rest of this paper is organized as follows. First, we present a brief overview of related works on age classification in Section 2. Then, in Section 3, we describe the collection and preprocessing of the dataset used in this study. Although this work is focused on automatic feature extraction, we show how manually selected features can be extracted using pose estimation in Section 4. Then, Section 5 describes how automatic feature extraction is achieved. Finally, experiments and results are discussed in Section 6, and conclusions and recommendations are given in Section 7.

---

[1] MMU GAG is available at https://github.com/timiderinola/mmu_gag.

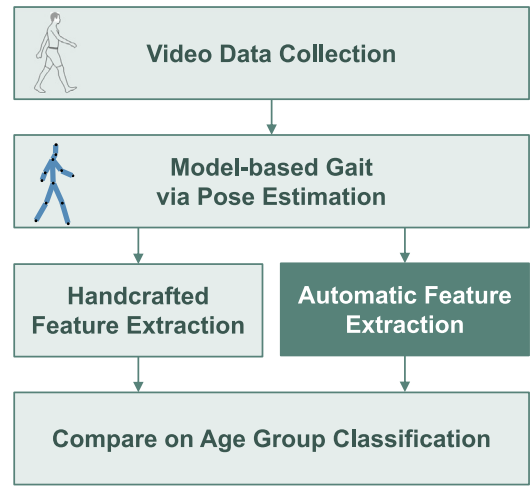

**Fig. 1.** Overall research process. First, we collected video data and then performed pose estimation to obtain the model-based gait of subjects. From the model-based gait, we compare the performance of auto-extracted features with manually extracted features in age group classification.

**Table 2**
Strengths and limitations of common descriptors for age group classification.

| Descriptor | Strengths | Limitations |
|---|---|---|
| Face (local) | Face shape and geometry contain age-associated information | Requires manual feature selection, which requires domain expertise. |
| Face (holistic) | Considers both geometry and texture. | Performance degrades due to poor image quality or occlusion. |
| Model-free gait | Lightweight and require little computational power. Most datasets are model-free. | Sensitive to carried objects and apparel. Requires background subtraction. |
| Model-based gait | Robust to carried objects and apparel. | Current techniques require manual feature selection. Few public datasets. |
| Sensor-based gait | Inertial sensors are low-cost, ubiquitous and precise. | Sensors must be carried by the subjects. |

## 2. Related works

In this section, we present a brief review of related works on age classification using gait and other biometric features. We also discuss some recent approaches that use gait graphs for person and action recognition using gait. We summarize the strengths and limitations of different descriptors used for age classification in Table 2.

### 2.1. Age classification and estimation problem using biometric features

The face is the most widely used biometric modality for age estimation and age group classification and has been deployed for public use (Burt, 2020). Face features are broadly classified as local or holistic. Holistic-based features are obtained from the raw pixel data of faces (Han, 2020; Liao et al., 2018; Cao et al., 2020; Akbari et al., 2021; Xie and Pun, 2020; Sharma et al., 2022; Dagher and Barbara, 2021). Local face features are obtained from the geometry of facial landmarks like eyes, mouths, and noses (Rahman et al., 2020; Rizwan et al., 2022). Some approaches to age group classification also use a fusion of local and holistic face features (Zhang et al., 2018). The main limitation of using face features is that performance depends on face image quality.

Other biometric features used for human age group classification include voice (Kwasny and Hemmerling, 2021; Si et al., 2022) and ear images (Moolla et al., 2021; Yaman et al., 2019). However, these modalities are limited in application as data is not easily obtained in unconstrained environments. For example, voice signals in the wild are either missing or distorted by noise, face or ear images in the wild could be partially or fully occluded.

### 2.2. Age prediction using gait features

Age-associated gait features can be obtained using wearable sensors (sensor-based) or from videos (vision-based). The vision-based approach could be further sub-classified as model-free or model-based. Model-free gait features are based on subjects' shape and appearance, and extracting them involves background subtraction and silhouette generation. The most common model-free gait descriptor is the Gait Energy Image (GEI) (Han and Bhanu, 2006). Since GEIs are represented as grayscale images, deep learning has been widely leveraged to extract age-associated features from GEIs (Zhu et al., 2020; Berksan, 2019; Lau and Chan, 2023; Russel and Selvaraj, 2021; Kitchat et al., 2022; Song et al., 2021; Zhang et al., 2022). In addition, some other works perform manual feature extraction from GEIs for age estimation or age group classification (Mansouri et al., 2018; Nabila et al., 2018). However, since obtaining silhouette images requires accurate background subtraction, obtaining model-free gait age features in the wild is infeasible.

Unlike model-free gait, model-based gait age features such as length of body segments, body ratios, and joint angles/range of motion can be manually extracted in the wild. These features have been classified as biological, kinematic, or hybrid (Aderinola et al., 2021b). Biological features are obtained from measuring body parts, such as limb length and head-to-body ratio. While biological features are sufficient to differentiate children from adults (Aderinola et al., 2021a), more fine-grained age group classification requires kinematic features such as gait speed and joint angles (Begg et al., 2005; Yang and Wang, 2015; Zhang et al., 2010). However, several model-based approaches to age group classification use biological and kinematic features (Aderinola et al., 2021a; Hema and Pitta, 2019; Punyani et al., 2018; Yoo and Kwon, 2017). The main limitation of these approaches is that manual feature extraction is tedious and requires domain expertise.

### 2.3. Gait graphs for person and action recognition

Until recently, model-based gait had been suitable for conventional machine learning techniques only. However, with the recent advances in human pose estimation, it is now possible to obtain joint coordinates automatically and with high accuracy. Hence, pose-based gait analysis is receiving increasing attention. Certain studies, such as (Liao et al., 2017), have leveraged Deep Neural Networks (DNNs) for gait recognition on model-based gait features obtained via state-of-the-art pose estimation algorithms like AlphaPose (Fang et al., 2022). Recent works such as Li et al. (2020), Liao et al. (2020), Teepe et al. (2021) have automatically leveraged graph neural networks to learn gait and action recognition features. However, they are not transferred for age group classification problems.

## 3. Data collection

In this section, we describe the video data collection and preprocessing steps. We also describe how motion data is obtained from the videos via pose estimation.

### 3.1. Dataset description

The gait in the wild (GITW) dataset contains video recordings of healthy individuals walking at a normal pace in unconstrained environments. It was collected from public domain online videos (see Fig. 2). It contains 439 subjects with one clip per subject. Since the videos were collected online, the views and covariate conditions are unconstrained. The subjects were labeled with the apparent gender and age group — child, adult, or senior. The dataset has 155 children, 177 adults, and 107 seniors.

### 3.2. Data preparation

Fig. 3 shows the overall process followed in data preparation. Video pre-processing, pose estimation, and coordinate smoothing are discussed in this section, while body graph construction and body parts aggregation are discussed in Section 5.2.

#### 3.2.1. Video preprocessing

As shown in Fig. 4, GITW videos are segmented manually using a linear video editor[2] to ensure only one subject in each video. In addition, each video is segmented based on the walking direction of the participants — left, right, front, or back.

#### 3.2.2. Pose estimation

Pose estimation is performed using AlphaPose (Fang et al., 2022) on a Windows 10 desktop computer with 8 GB of RAM, 8 GB GPU, and a core i5 Intel processor. For each subject with $J$ body keypoints in a video with $F$ frames, the pose estimation algorithm outputs a sequence of tuples $\{(x_i^t, y_i^t, c_i^t) | i = 1, \ldots, J; t = 1, \ldots, F\}$, where $(x_i^t, y_i^t)$ represent the body keypoints coordinates, and $c_i^t = [0, 1]$ is the confidence score for joint $i$ in frame $t$.

**Data Augmentation.** Since videos recorded by participants are of arbitrary lengths, $F$ is standardized to 50 frames. Hence, sequences longer than 50 frames were divided into separate frames. However, clipped sequences shorter than 25 frames were discarded as too short. For further augmentation, pose sequences were reversed, simulating the backward movement of subjects.

#### 3.2.3. Coordinates smoothing

The pose estimation output can be regarded as a multivariate time series, requiring some smoothing. However, smoothing with conventional filters could result in losing relevant information. Hence, we propose z-score smoothing, which replaces sudden spikes in the coordinate values with the average of the two adjacent coordinate values. Z-score smoothing works on the assumption that the motion time series data of subjects walking at a normal pace will have no sudden spikes. In the absence of sudden spikes, the data is left unchanged. After z-score smoothing, pairs of adjacent frames are averaged for dimensionality reduction, such that a sequence containing 50 frames is reduced to 25.

Let $\mu$ be the mean of $\boldsymbol{d} \in \mathbb{R}^F$, where $\boldsymbol{d}$ is the univariate movement time series for a body keypoint across $F$ frames. The frame-to-frame standard deviation is:

$$\sigma^2 = (1/F) \sum_{i=1}^{F}(d_i - \mu)^2, \ \sigma^2 \in \mathbb{R} \tag{1}$$

---

[2] Wondershare Filmora 9, https://filmora.wondershare.com

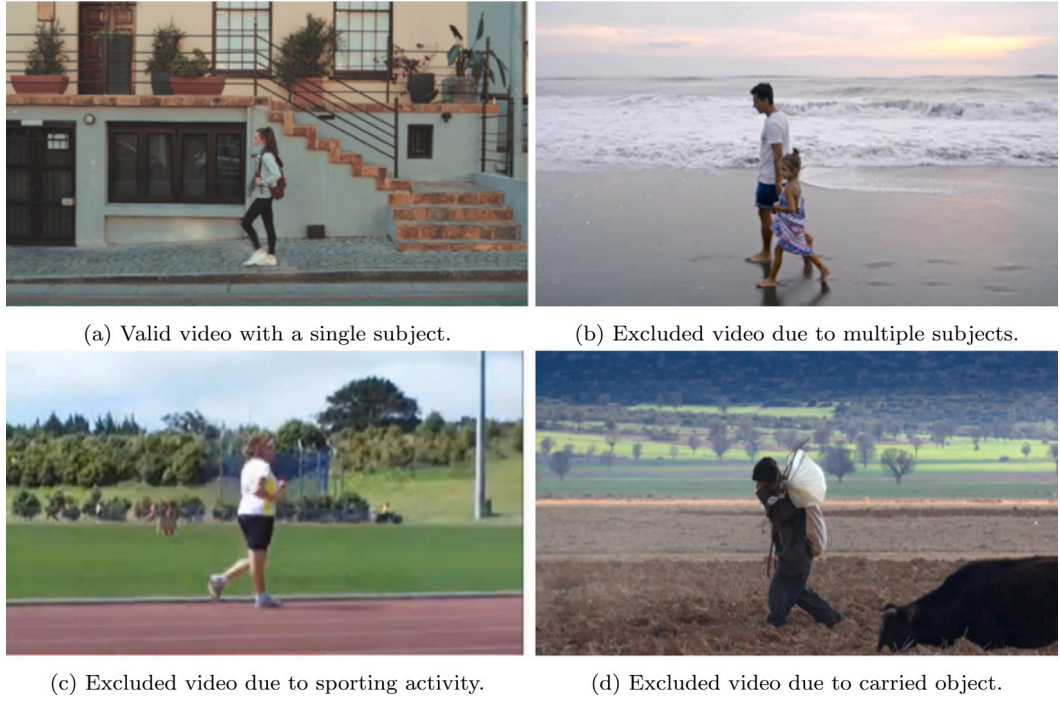

(a) Valid video with a single subject.

(b) Excluded video due to multiple subjects.

(c) Excluded video due to sporting activity.

(d) Excluded video due to carried object.

**Fig. 2.** GITW data collection videos. Videos with pathological gait and those involving sporting activities such as running and jogging are excluded. In addition, except for seniors with walking sticks, videos of subjects with carried objects are also excluded.

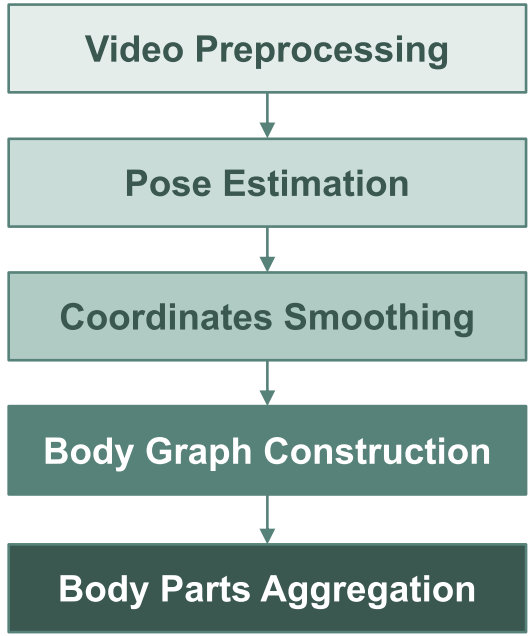

**Fig. 3.** Data Preparation Steps. After preprocessing, we obtained body keypoint coordinates via pose estimation. We then performed smoothing and obtained body graphs and part graphs from the keypoint coordinates.

The absolute z-scores for each element in the sequence are obtained as

$$z_i = |(d_i - \mu)/\sigma^2|, \; i = 1, \ldots, F; \; \mathbf{z} \in \mathbb{R}^F \qquad (2)$$

A smoothened sequence $\mathbf{d}_s \in \mathbb{R}^F$ is then obtained according to the rule:

$$\delta_{s_i} = \begin{cases} d_i, & z_i \leq \tau, \\ (d_{i-1} + d_{i+1})/2, & \text{otherwise} \end{cases} \qquad (3)$$

where $\tau$ is a threshold value defined as $\tau = (1/F) \sum_{i=1}^{F} z_i$. The pseudocode algorithm for z-score smoothing is given in Algorithm 1. Fig. 5 illustrates the effects of z-score smoothing on a very noisy neck keypoint sequence.

---

**Algorithm 1:** Pseudocode Procedure for Z-Score Smoothing

**Data:** $d$
**Result:** $\mathbf{d}_s$
1 initialization;
2 $\mu \leftarrow MEAN(d)$;
3 $std \leftarrow STANDARDDEVIATION(d)$;
4 $z \leftarrow ABS((d - \mu)/std)$;
5 $threshold \leftarrow MEAN(z)$;
6 INITIALIZE i = 0; list $\mathbf{d}_s$;
7 **while** $i < LENGTH(\mathbf{d})$ **do**
8     **if** $d_i >$ threshold **then**
9        $d_{s_i} \leftarrow MEAN(d_{i+1}, d_{i-1})$
10     **else**
11        $d_{s_i} \leftarrow d_i$;
12     **end**
13     INCREMENT $i$;
14 **end**
15 RETURN $\mathbf{d}_s$;

---

## 4. Handcrafted gender and age-associated features extraction

In this section, we extract handcrafted gender and age-associated gait features from the GITW dataset for analysis and comparison with automatically extracted features. Let $u = \{x_i^t, y_i^t\}$ and $v = \{x_{i+1}^t, y_{i+1}^t\}$ be the coordinates of two adjacent keypoints in frame $t$, the Euclidean distance, $D_E$ between $u$ and $v$ is given by:

$$D_E(u, v) = [(x_i^t - x_{i+1}^t)^2 + (y_i^t - y_{i+1}^t)^2]^{1/2} \qquad (4)$$

Hence, as shown in Table 3, subjects' gait features are obtained from the keypoint coordinates obtained via pose estimation. The statistics of

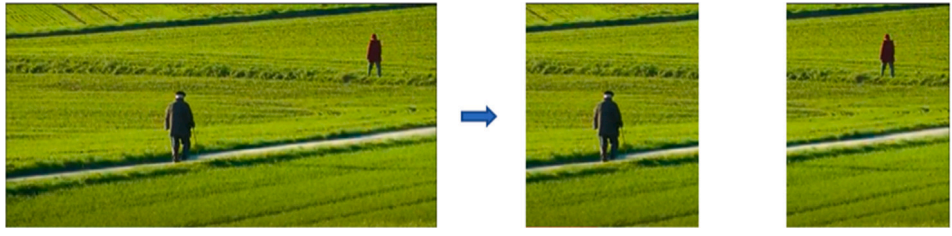

**Fig. 4.** Video with Multiple Subjects (Left). Cropped Videos (Right). GITW videos were segmented to get the clips of interest. For example, if the subject appears only in the last 20 s, all the other parts of the video are discarded. Videos with multiple subjects were cropped into different videos, ensuring that each had no more than one subject.

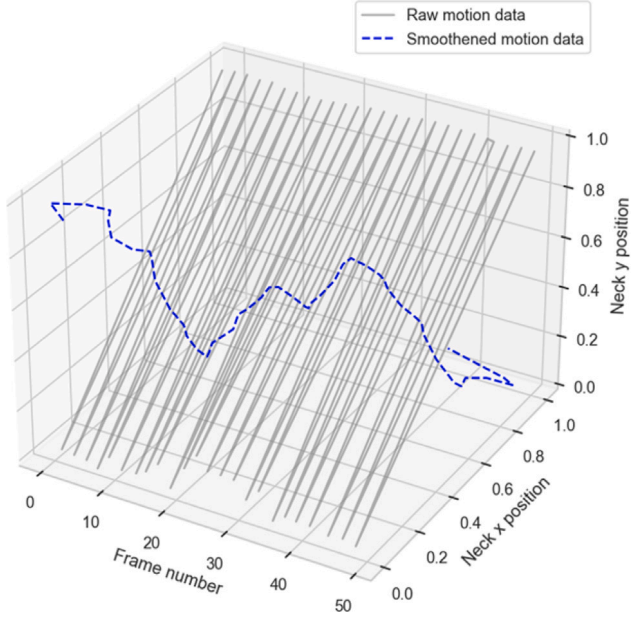

**Fig. 5.** Coordinate Z-score Smoothing on a Noisy Sample. The sample shown in this illustration is very noisy due to false neck detections during pose estimation. These false detections result in spikes in the time series of neck motion. Z-score smoothing works on the assumption that the motion time series data of subjects walking at a normal pace will have no spikes.

**Table 3**
Definition of features extracted from body keypoints.

| Feature (length) | Description |
| --- | --- |
| *head* (head length) | $D_E(head, neck)$ |
| *hand* (Upper limb length) | $D_E(shoulder, elbow) + D_E(elbow, wrist)$ |
| $body_u$ (Upper body length) | $D_E(neck, hip)$ |
| *leg* (Lower limb length) | $D_E(hip, knee) + D_E(knee, ankle)$ |
| *stature* (Total body length) | $head + body_u + leg$ |
| $hand_g$ (Hand-to-ground) | $body_u + leg - hand$ |
| *step* (Step length) | $D_E(ankle_{left}, ankle_{right})$ |
| *cadence* (steps per minute) | $s/N$ |
| *speed* (gait cycles per second) | $step \times cadence$ |

Estimating age-associated features from pixel coordinates of *head*, *neck*, *shoulder*, *elbow*, *wrist*, *hip*, *knee*, and *ankle*. Except for the *ankle*, The side of the keypoints closest to the camera (left or right) is chosen.

handcrafted features obtained from GITW are summarized in Table 4, and boxplots are given in Fig. 6.

## 5. Automatic age feature extraction

In this section, we discuss how we extract features automatically from gait graphs. First, we discuss how we construct spatiotemporal gait graphs. Then, we discuss the building blocks of the Part-Adaptive Residual Graph Convolutional Neural Network.

### 5.1. Preliminary: Body graph construction

From the output of pose estimation described in Section 3.2.2, the body is modeled as a connected graph, $body = \{joints, bones\}$, where the detected joints are the nodes (the black dots in Fig. 7), and the bones are edges. First, we construct the body graph for a single frame. Then, using the constructed body graph, we construct a spatiotemporal body graph to represent the motion across all frames.

We represent the body graph in a single frame structurally with an adjacency matrix $A_{body} \in \mathbb{R}^{J \times J}$, and feature-wise with $X_{body} \in \mathbb{R}^{J \times 3}$, where $J = |joints|$.

The spatiotemporal body graph across all frames consists of several body graphs, and it has spatial edges ($E_{spatial}$) and temporal edges ($E_{temporal}$). $E_{spatial}$ is the set of all edges of the constituent body graphs. $E_{temporal}$ consists of the connections between nodes of the constituent body graphs to their counterparts in adjacent frames. An example of $E_{temporal}$ is the connection between the right ankle in frame 1 and the right ankle in frame 2. Hence, we represent the walking sequence of each subject across $F$ frames as a spatiotemporal graph $G = \{V, E\}$. $V = \{v_{i,j} | i = 1, \ldots, F; j = 1, \ldots, J\}$ is the set of all keypoints detected across all video frames for a subject, while $E = \{E_{spatial}, E_{temporal}\}$.

Suppose $K = |E_{spatial}|$ is the total number of bones detected across $N$ frames, and $L = |E_{temporal}|$. Since each pair of adjacent frames has $J$ temporal connections — one for each joint, hence $L = J(F-2)$ and the total number of edges, $|E| = K + L$. Each spatiotemporal body graph has a node feature matrix $X \in \mathbb{R}^{|V| \times 3}$, where $|V|$ is the number of nodes across all video frames. Structurally, the body graphs are represented by a square adjacency matrix $A \in \mathbb{Z}^{|E| \times |E|}$ that captures both spatial and local connections of body keypoint coordinates. The spatiotemporal body graph is illustrated in Fig. 8.

### 5.2. Body parts aggregation

Human motion is not regarded as a movement of joint collections but as an interdependent movement of body parts, which we call aggregation of joints. For instance, with reference to Table 3, the ***spine*** is an essential determinant of upper body posture, consisting of the head, neck, and central hip. The ***arms***, consisting of the shoulder, elbow, and wrist, are vital to balance in walking, and the arm's motion can discriminate subjects as a child, adult, or senior (Fig. 6). The body's center of mass is at the ***hip***, which comprises the left, central, and right hips. The ***legs*** are arguably the most critical parts of locomotion. Besides being the center of pressure, dynamic features such as gait speed, step length, and cadence are determined by leg movement, and these are crucial in more fine-grained classification. Therefore, we propose a part aggregation scheme that groups the body graph into four parts: *spine*, *arm*, *hip*, and *leg*, each made up of three joints.

As illustrated in Fig. 9, each part in the part graph is associated with four features based on the raw coordinates aggregation: (a) range of motion (ROM) $\theta$ at the central node, (b) total length $L$ of the part, (c) swing $S$, the distance between the left and right sides, and (d) distance $D$, measured as the distance in pixels covered per frame. Suppose $b$ is

**Table 4**
Gait gender and age-associated handcrafted features extracted from GITW dataset.

| Feature | Female | | | Male | | |
|---|---|---|---|---|---|---|
| | *child* #=70 | *adult* #=77 | *senior* #=44 | *child* #=85 | *adult* #=100 | *senior* #=63 |
| HB | 0.20 ± 0.05 | 0.15 ± 0.02 | 0.15 ± 0.02 | 0.20 ± 0.04 | 0.14 ± 0.02 | 0.14 ± 0.03 |
| HG | 0.50 ± 0.06 | 0.5 ± 0.04 | 0.52 ± 0.04 | 0.49 ± 0.05 | 0.53 ± 0.03 | 0.53 ± 0.05 |
| LG | 0.42 ± 0.06 | 0.49 ± 0.04 | 0.46 ± 0.03 | 0.41 ± 0.05 | 0.48 ± 0.03 | 0.48 ± 0.03 |
| SL | 0.43 ± 0.15 | 0.33 ± 0.11 | 0.29 ± 0.09 | 0.44 ± 0.13 | 0.37 ± 0.10 | 0.32 ± 0.10 |
| CD | 52.59 ± 24.46 | 54.08 ± 24.65 | 47.01 ± 25.91 | 53.75 ± 26.13 | 65.2 4 ± 28.48 | 45.69 ± 28.12 |
| GS | 23.07 ± 13.02 | 19.14 ± 12.38 | 14.39 ± 11.38 | 24.43 ± 13.89 | 25.66 ± 15.02 | 15.76 ± 13.52 |

#: number of subjects; HB: Head-to-body ratio; HG: hand-to-ground distance; LG: lower limb length; SL: step length; CD: cadence (steps per minute); GS: gait speed (cycles per minute). Mean values are given ± standard deviation.

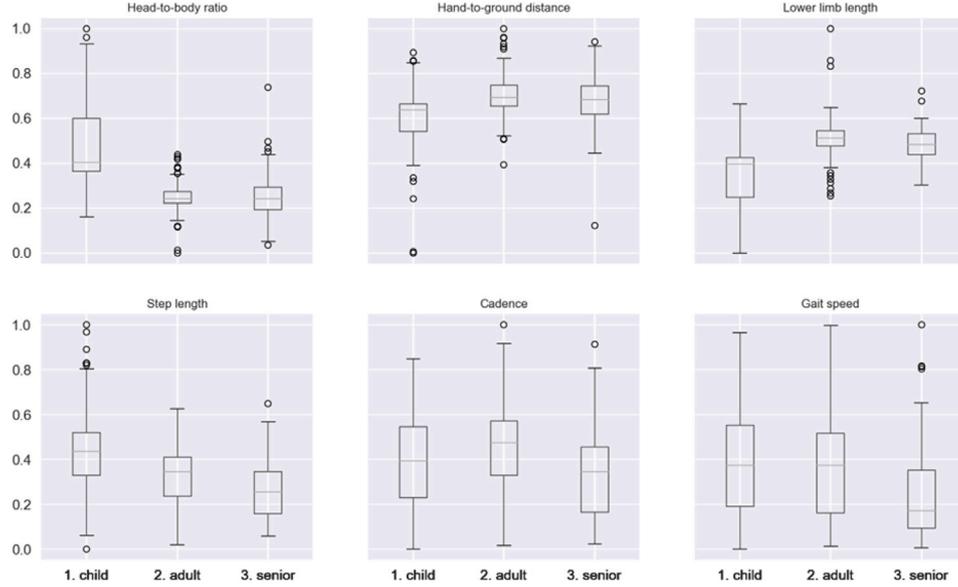

**Fig. 6.** GITW Subjects' Average Features by Age Group. Children in the GITW dataset are distinguishable by their greater head-to-body ratios and have a smaller hand-to-ground distance than the other classes. The smaller hand-to-ground distance of children follows directly from the shortness of their lower limbs compared to adults and seniors. Seniors in the dataset have the slowest gait speed.

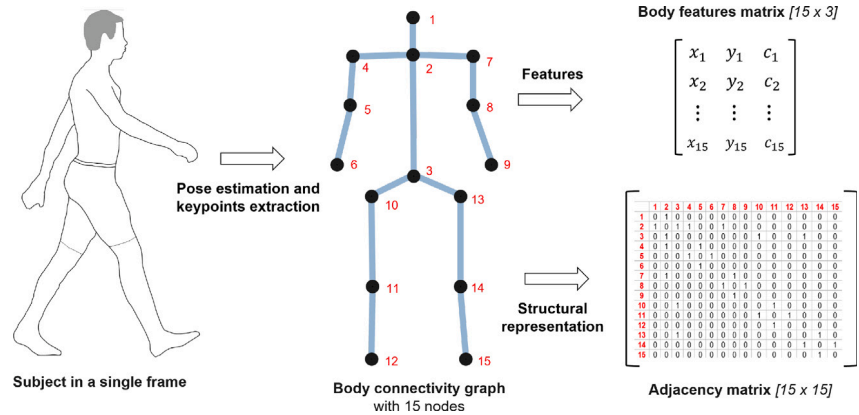

**Fig. 7.** Body graph construction in a single frame. Each keypoint is represented with a row in the body features matrix while the body structure is represented with an adjacency matrix.

the central node with $a$ and $c$ as its adjacent nodes. From $\vec{ba}$ and $\vec{bc}$, $\theta$ and $L$ are obtained as:

$$L = |\vec{ba}| + |\vec{bc}| \qquad (5)$$

$$\theta = \cos^{-1}((ba \cdot bc)/|ba||bc|) \qquad (6)$$

Suppose that in frame $i$, the central node $b_i$ for a part is at coordinate $(x_{b_i}, y_{b_i})$ in the 2D plane, D is obtained as the Euclidean distance between $b_i$ and $b_{i+1}$:

$$D(b_i, b_{i+1}) = [(x_{b_i} - x_{b_{i+1}})^2 + (y_{b_i} - y_{b_{i+1}})^2]^{1/2} \qquad (7)$$

Each part has two extremes: *head* and *mid-hip* for the *spine*; *left* and *right* each for the *arm*, *leg*, and *hip*. Let $p_i$ and $q_i$ be the extremes for an arbitrary part in frame $i$. The swing $S_i$ for each frame part is defined as:

$$S_i(p_i, q_i) = [(x_{p_i} - x_{q_i})^2 + (y_{p_i} - y_{q_i})^2]^{1/2} \qquad (8)$$

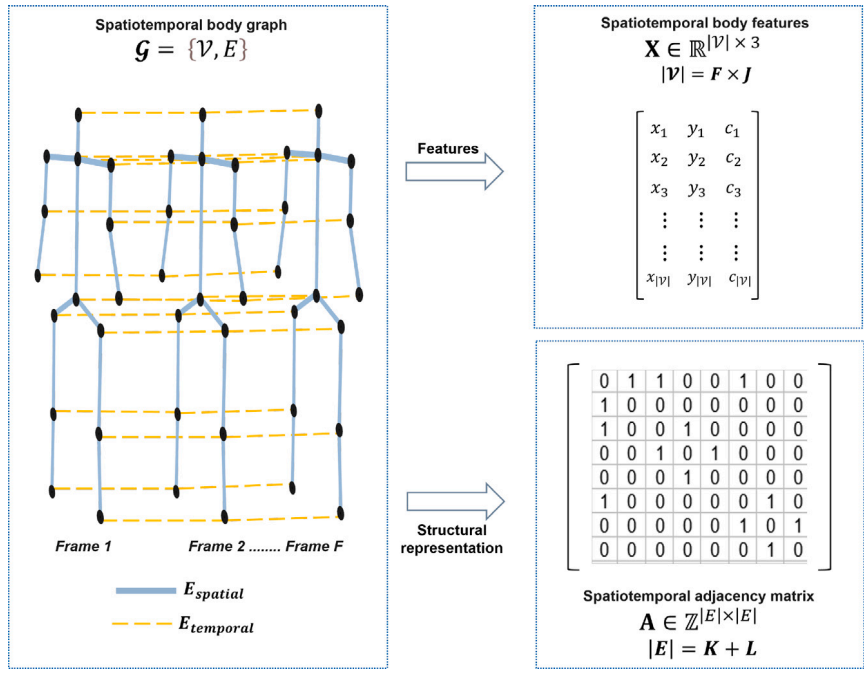

**Fig. 8.** Spatiotemporal Body Graph Representation. Each spatiotemporal graph represents the walking motion of a single subject in a video.

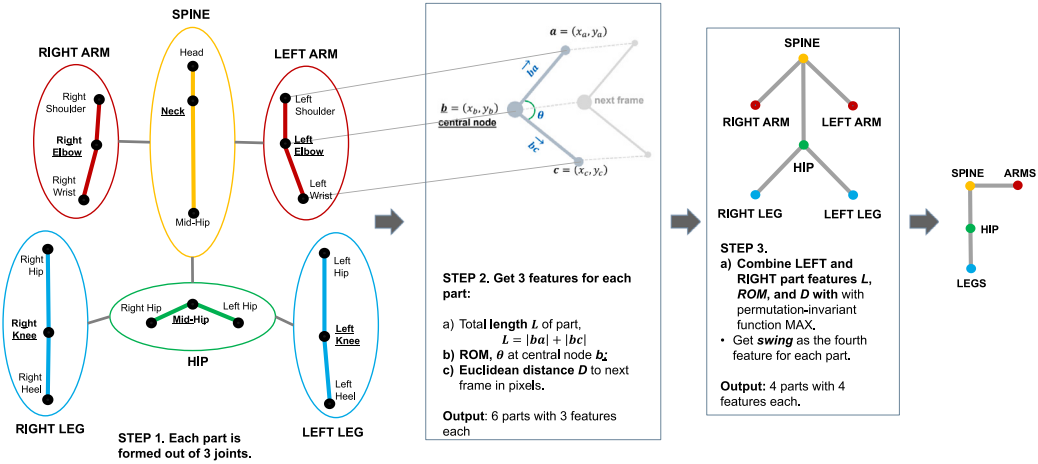

**Fig. 9.** The Three Steps of Body Parts Aggregation. Each part consists of three nodes in the body graph aggregated to form a part node. The central node in each part is underlined and shown in bold.

Hence, the part graph is defined as a spatiotemporal graph $G_{part} = \{V_{part}, E_{part}\}$, as described in Fig. 10. Each part graph has a node feature matrix $X_{part} \in \mathbb{R}^{|V_{part}| \times 4}$, where $|V_{part}| = 4F$ represents the part nodes across all video frames, each with four features described above. $E_{part}$ is the set of all spatial edges and temporal edges. Hence, with three spatial edges per frame and four part nodes, $|E_{part}| = 3 + 4(F-2)$. Structurally, the part graphs are represented by a square adjacency matrix $A_{part} \in \mathbb{Z}^{|E_{part}| \times |E_{part}|}$.

### 5.3. Part-adaptive residual graph convolutional neural network

#### 5.3.1. Preliminary

This section presents the Part-Adaptive Residual Graph Convolutional Neural Network (PairGCN) with adaptive edge weights. PairGCN takes part graphs (or body graphs) as input and learns discriminative features for children, adults, and seniors. PairGCN uses gait features only and includes an enhancement that learns edge weights for the part graph adjacency matrix in the first layer. The proposed PairGCN is based on Morris et al.'s GCN operator (Morris et al., 2019), given as:

$$x_i^{l+1} = \theta_1 x_i^l + \theta_2 \sum_{j \in \mathcal{N}(i)} w_{j,i} \cdot x_j^l \quad (9)$$

where, $x_i^l$ denotes the embedding of node $i$ (target node) in layer $l$; $\mathcal{N}(i)$ is the set of neighbors of node $i$; $x_j^l$ denotes the embedding of node $j$ (source node) in layer l; $\theta_1$ and $\theta_2$ are learnable parameters; and $w_{j,i}$ is the edge weight from source node $j$ to target node $i$.

In conventional GCN, $w_{j,i}$ is often set to a constant value for all edges. This often suffices in domains where the main discriminative features of the graph are the node features and graph structure. Examples include action recognition and gait recognition (Li et al., 2020; Teepe et al., 2021). Although extracting age-associated features from a body skeleton graph also requires the body structure and joint positions represented as node features, the connectivity of the human body contains no age-associated information. The weights associated with bones and body parts and other dynamic information such as range of motion

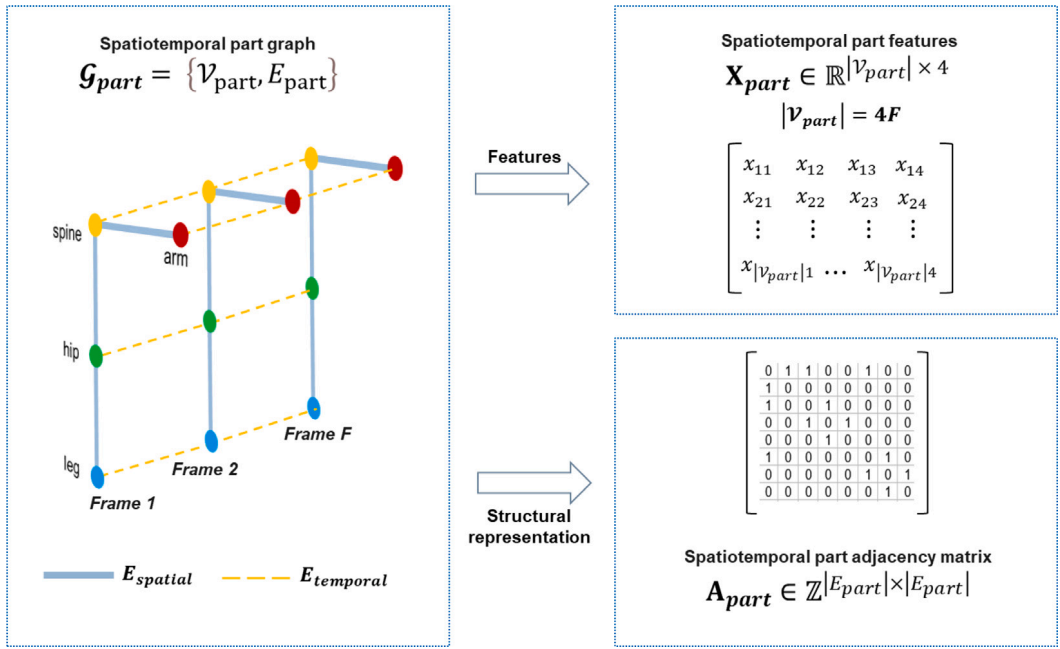

**Fig. 10.** Spatiotemporal Parts Graph Representation. The spatial connections are represented using spatial edges, $E_{spatial}$, while the inter-frame connections are represented using temporal edges, $E_{temporal}$.

and speed must also be adequately modeled to extract age-associated features.

Therefore, the proposed PairGCN is modified from the GCN operator in Eq. (9) by introducing an adaptive edge weight $\theta_w \in \mathbb{R}^{|E_{part}|}$, such that

$$x_i^{l+1} = \theta_1 x_i^l + \theta_2 \sum_{j \in \mathcal{N}(i)} \theta_{w_{j,i}} \cdot x_j^l \qquad (10)$$

How PairGCN captures age-associated features based on body parts is discussed in Section 6.1.

*5.3.2. Part-adaptive GCN*

As illustrated in Fig. 11, the first layer of PairGCN performs the essential GCN operation according to Eq. (9). For the first layer, it makes use of a constant $\theta_w$, initialized as 1 for each edge, and node embeddings are obtained for each graph as an aggregation of node features and their 1-hop neighborhood. Then, after the first layer, $\theta_w$ is updated based on the node embeddings obtained in the first layer as

$$\theta_w' = MEANPOOL(GCN(x^l))\theta_w \qquad (11)$$

where $MEANPOOL(\cdot)$ denotes the graph-wise mean of the first layer node embeddings, $GCN(\cdot)$ refers to the forward pass performed by GCN and $x^l$ is the node embedding in layer $l, l = 1, \dots, L$. Since edges are formed with 1-hop neighbors, edge weights $\theta_w$ are updated based on first layer node embeddings. Subsequent layers then make use of the updated edge weight in obtaining node embeddings by operating:

$$x_i^{l+1} = \theta_1 x_i^l + \theta_2 \sum_{j \in \mathcal{N}(i)} \theta_{w_{j,i}}' \cdot x_j^l \qquad (12)$$

The overall architecture of PairGCN with $N$ residual blocks is given in Fig. 12. With an added fully connected layer, PairGCN is trained with the guidance of age group labels.

*5.3.3. Loss function*

Data preparation steps such as augmentation and segmentation based on view introduce class imbalance in the training dataset. In PairGCN, two approaches are used to solve the class imbalance problem. Firstly, weighted random subsampling is used, such that underrepresented classes are oversampled in each training batch. Secondly, PairGCN uses a class-balanced cross-entropy loss that weights the total loss based on each class's effective number of samples. To get the weighted losses, class weights are obtained based on the effective number of samples (Cui et al., 2019) per class which is given as

$$E_c = (1 - \beta^c)/(1 - \beta) \qquad (13)$$

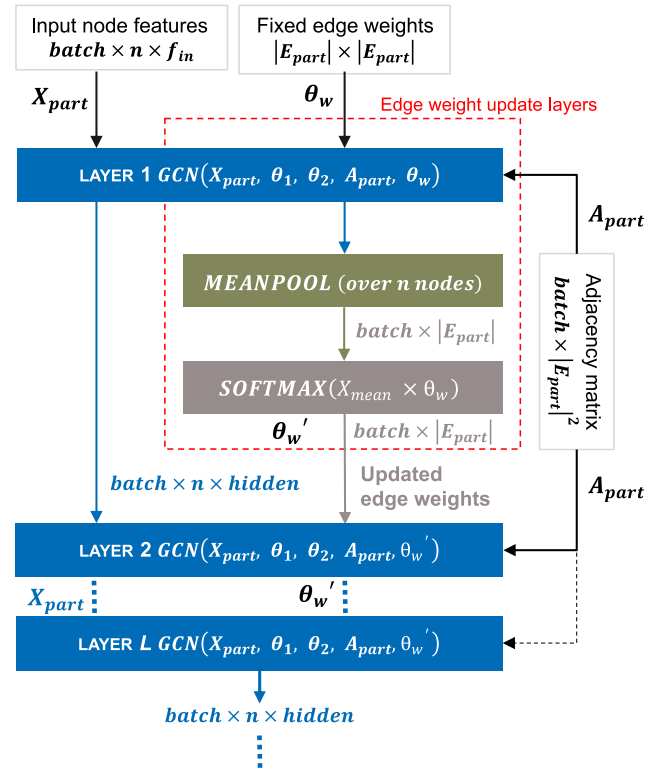

**Fig. 11.** Updating Edge Weights in PairGCN. $batch$: batch size; $n$: number of nodes; $f_{in}$: number of input features; $hidden$: number of hidden neurons.

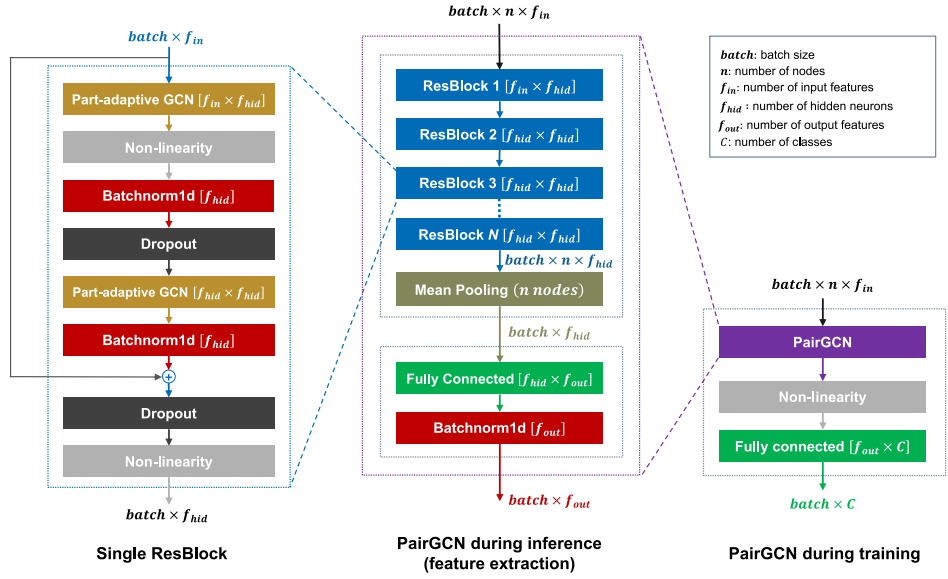

**Fig. 12.** PairGCN Pipeline. During training, PairGCN is trained with age group labels. During inference, PairGCN is used as a feature extractor.

where $c$ = total number of samples of a class; $\beta = (N-1)/N$; $N$ = total number of samples. Then, the class-balanced loss $L_{cb}$ is defined as:

$$L_{cb}(p, y) = 1/E_{c_y}(L(p, y)) \quad (14)$$

where $y$ is the class label and $p$ is the softmax probability. For this use case, $L(p, y)$ is the cross-entropy loss function for $C$ classes given as:

$$L_{cb}(p, y) = (1-\beta)/(1-\beta^{c_y}) \sum_{i=1}^{C} y_i \log p_i \quad (15)$$

#### 5.3.4. Inference

During inference, PairGCN is used as a feature extractor for unlabeled body or part graphs as input, for which it outputs a feature vector that describes the age-associated gait features in the graph. Due to age's ordinal and hierarchical nature, features obtained from PairGCN are best suited for hierarchical age group classification with tree-based classifiers such as Random Forest. However, any classifier could be fitted on the gait embedding obtained from PairGCN and used for final classification.

### 6. Experiments and results

In this paper, GITW is used to evaluate the proposed method since it contains samples for *child*, *adult*, and *senior* classes. The dataset is divided into two main views with respect to the camera, namely, the front view (0°–45°) and the side view (45°–90°). Training and inference were performed on a Windows 11 PC with 8 GB RAM, 2 GB GPU, and a core i7 Intel processor. During the training phase, each network is trained with the guidance of the age groups. In the inference stage, the trained model is used as a feature extractor to obtain the gait age signatures. For each subject, the model outputs a feature vector $f_{out} \in \mathbb{R}^{64}$, which is used as the gait age signature for age group classification. The list of variables and hyperparameters are shown in Table 5.

A Random Forest classifier with 150 trees is used for age group classification. In addition to classification accuracy, the class F1 Scores are used as evaluation metrics. Each layer of a Graph Neural Network performs an aggregation of features of a node and its neighbors. Therefore, the depth of the neural network should not greatly exceed the maximum depth of the graph being considered. Hence, all experiments were conducted using three residual blocks, a learning rate of 0.001, a batch size of 64, and run for 200 epochs. The GITW dataset was divided into the train, validation, and test sets. The train set was 75% of the dataset, and the validation set was 25% of the train set (Table 6).

Table 7 shows that the side view of gait generally offers more age-associated features, especially for children. In comparison, the front view of gait offers more age-associated features for adults and seniors, as shown by the higher F1-score for both classes in the (45°–90°) view. For all views, using Random Forest gives better results.

#### 6.1. An analysis of adaptive weight distributions and body parts

The spatiotemporal part graph obtained via parts aggregation (Fig. 9) consists of spatial edges $E_{spatial}$ and temporal edges $E_{temporal}$. The spatial edges are SPINE–ARM, SPINE–HIP, and HIP–LEG, while the temporal edges are SPINE–SPINE, ARM–ARM, HIP–HIP, and LEG–LEG. For each edge in the part graph, the edge formed by part $j$ and $i$ is weighted as $\theta'_{w_{j,i}} \in [0, 1]$. These weights are averaged for each age group (Fig. 13).

The SPINE–ARM and ARM–ARM connections are important in discriminating between children and other age groups. This is as expected since children are easily distinguishable by their arm swings. However, the arm movement of seniors is often more cautious since they are prone to falls. On the other hand, SPINE–HIP and HIP–LEG connections are assigned greater weights in distinguishing seniors. This follows since seniors have a greater range of motion at the hips and lower extremity joints. The equally weighted LEG–LEG connection indicates the universal importance of dynamic features such as gait speed, step length, and cadence, which depend on the temporal movement of the lower limbs.

#### 6.2. Ablation studies

This section demonstrates the ablation studies on the performance of PairGCN architecture, adaptive edge weights, body parts aggregation, coordinate smoothing, and view regarding age classification accuracy and F1 scores on the GITW dataset.

*A. PairGCN vs. Baseline GCN architecture*

GCN architectures are often highly task-specific. For comparison, a baseline GCN architecture (Fig. 14) is constructed according to the basic architecture proposed by You et al. (2020). As shown in Table 8, PairGCN achieves the highest overall accuracy on the GITW dataset in the 45°–90° view. As expected, the Baseline GCN and PairGCN both learn more age-associated features in the side view of gait. This is expected since most age-associated features such as arm swing, posture, step length, stride length, upper and lower limb range of motion, and minimum toe clearance are found in the side view of gait. The front view of gait presents mostly static features, while the side view is rich with both static and dynamic spatiotemporal gait features.

**Table 5**
List of variables and hyperparameters used in experiments.

| Variable/Description | | Values | | |
| --- | --- | --- | --- | --- |
| | | *BaselineGCN* | *SequenceGCN* | *PairGCN* |
| *batch* | Batch size | 64 | 64 | 64 |
| $C$ | Number of classes | 3 | 3 | 3 |
| $F$ | Total number of frames | 50 | 50 | 50 |
| $f_{in}$ | Number of input features | 3 | 3 | 4 |
| $f_{hid}$ | Number of hidden neurons | 64 | 64 | 64 |
| $f_{out}$ | Number of output features | 64 | 64 | 64 |
| $n$ | Number of nodes in input graph | 15 | 15 | 4 |
| $|E|$ | Spatiotemporal body graph edges | 1420 | N/A | N/A |
| $|E_b|$ | Bones/edges in a single frame | N/A | 14 | N/A |
| $|E_{part}|$ | Spatiotemporal part graph edges | N/A | N/A | 342 |
| $|E_{body}|$ | Sequence body graph edges | N/A | 700 | N/A |
| $N$ | Number of residual blocks | N/A | N/A | 3 |
| | Learning rate | 0.001 | 0.001 | 0.001 |
| | Training epochs | 200 | 200 | 200 |
| | Non-linearity | $LeakyReLU(negative\_slope = 0.01)$ | | |

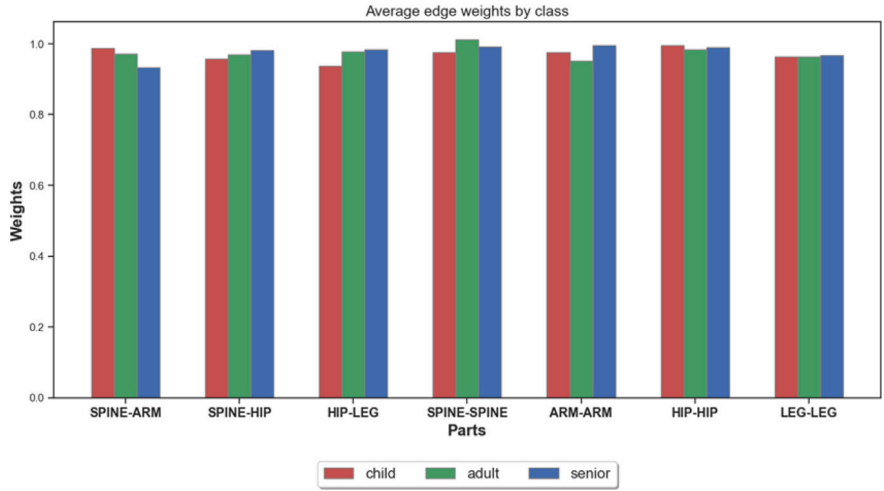

**Fig. 13.** PairGCN Average Edge Weights by Class, showing the average weights for the spatial edges: SPINE–ARM, SPINE–HIP, and HIP–LEG, and the temporal edges: SPINE–SPINE, ARM–ARM, HIP–HIP, and LEG–LEG.

**Table 6**
Experiment setup for the GITW dataset.

| View | Total Samples | After augmentation | Train | Val | Test |
| --- | --- | --- | --- | --- | --- |
| (0°–45°) | 96 | 984 | 552 | 186 | 246 |
| (45°–90°) | 224 | 1698 | 958 | 316 | 424 |

*B. Adaptive edge weighting*

The performance of PairGCN with adaptive edge weights is compared to its performance with constant edge weights on the 45°–90° view of the GITW dataset. PairGCN without adaptive edge weights shows better F1 Scores for children than the older age groups in Table 9. This suggests that it becomes increasingly difficult to discriminate against older age groups without adaptive edge weighting.

*C. Part aggregation vs. Raw coordinates*

The performance of part aggregation is compared to raw coordinates for age group classification, using the Baseline GCN and PairGCN on the 45°–90° view of the GITW dataset. The two methods are compared in model performance (Tables 8 and 9). Part aggregation improves the performance of both architectures.

*D. Coordinate smoothing*

As shown in Table 9, the F1 scores suggest that without smoothing, PairGCN finds it increasingly difficult to learn features associated with age groups as the age increases. While it is straightforward to learn features associated with children (such as their body size) even with noisy coordinates, the features associated with adults and seniors are more subtle and sensitive to noise.

*6.3. The importance of temporal gait information*

In sequential data and time-series analysis, recurrent neural networks (RNNs) such as Long-Short Term Memory (LSTM) (Hochreiter and Schmidhuber, 1997) have achieved outstanding results in sequential classification and prediction. While it may seem more natural to model gait as a sequence and adopt RNNs, this section presents a comparative analysis between spatiotemporal and sequence-based gait for age group classification. A Sequence GCN with a Graph Gated Recurrent Unit (GraphGRU) architecture is constructed for this analysis, as illustrated in Fig. 15.

Sequence gait graphs are obtained where each subject is represented by a sequence of body graphs, with one body graph for each input video frame. Each body graph consists of joints as vertices, and bones as edges, defined as $G_b = \{V_b, E_b\}, V_b = \{V_{b_i} | i = 1, \ldots, J\}; E_b = \{E_{b_j} | j = 1, \ldots, |E_{body}|\}$. Structurally, the body graphs are represented by an identical square adjacency matrix $A_b \in \mathbb{Z}^{|E_b| \times |E_b|}$, where $|E_b|$ is the number of edges. Hence, for $F$ frames, each subject's gait sequence

**Table 7**
Summary of PairGCN results on GITW dataset.

| View | Classifier | F1 scores | | | Accuracy |
|---|---|---|---|---|---|
| | | child | adult | senior | |
| Front view (0°–45°) | PairGCN FC layer | 0.95 | 0.83 | 0.88 | 88% |
| | Random Forest | 0.95 | 0.96 | 0.97 | 96% |
| Side view (45°–90°) | PairGCN FC layer | 0.95 | 0.90 | 0.82 | 90% |
| | Random Forest | 0.99 | 0.99 | 0.99 | 99% |

FC layer: fully connected layer.

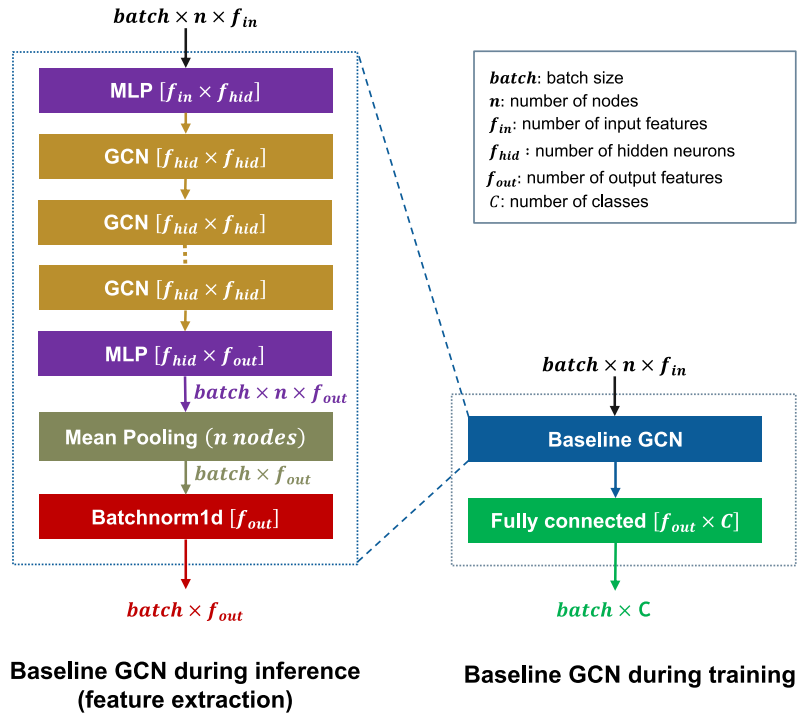

**Fig. 14.** Baseline GCN Architecture. During training, Baseline GCN is trained using age group labels. During inference, it is used as a feature extractor.

**Table 8**
Effects of different settings on accuracy.

| | Settings | | | | | Accuracy |
|---|---|---|---|---|---|---|
| | Arch. | Adaptive weights | Aggr. Features | Smoothing | View | |
| 1. | | ✗ | ✗ | ✓ | 45°–90° | 0.94 |
| 2. | BaselineGCN | ✗ | ✓ | ✓ | 45°–90° | 0.96 |
| 3. | | ✗ | ✓ | ✓ | 0°–45° | 0.97 |
| 4. | | ✓ | ✓ | ✓ | 0°–45° | 0.96 |
| 5. | PairGCN | ✗ | ✓ | ✓ | 45°–90° | 0.98 |
| 6. | | ✓ | ✗ | ✓ | 45°–90° | 0.98 |
| 7. | | ✓ | ✓ | ✓ | 45°–90° | 0.99 |

**Table 9**
Effects of different settings on F1 scores.

| | Settings | | | | F1 scores | | |
|---|---|---|---|---|---|---|---|
| | Arch. | Adaptive weights | Aggr. features | Smoothing | child | adult | senior |
| 1. | BaselineGCN | ✗ | ✗ | ✓ | 0.98 | 0.94 | 0.89 |
| 2. | | ✗ | ✓ | ✓ | 0.98 | 0.96 | 0.95 |
| 3. | | ✗ | ✓ | ✓ | 0.99 | 0.98 | 0.97 |
| 4. | PairGCN | ✓ | ✗ | ✓ | 0.99 | 0.98 | 0.96 |
| 5. | | ✓ | ✓ | ✗ | 1.00 | 0.99 | 0.97 |
| 6. | | ✓ | ✓ | ✓ | 0.99 | 0.99 | 0.99 |

is represented as a sequence of body graphs, $G_{seq} = \{G_{b_1}, \ldots, G_{b_F}\}$. SequenceGCN is trained using $G_{seq}$ and $A_b$ with the guidance of age group labels.

The results obtained by SequenceGCN (Table 10) suggest that the network cannot learn age-associated features from gait modeled as a sequence. Several reasons could be given for this observation. First, although gait is observed from the walking sequence of subjects, gait is more of a cycle than a sequence. However, sequence-based methods such as SequenceGCN will consider each frame in the context of previous frames while failing to consider the temporal connections between the frames. Hence, modeling gait as a spatiotemporal graph is more suitable for age group classification. In contrast, modeling gait as a sequence of body graphs is perhaps more suited to naturally sequence-based tasks such as human action recognition.

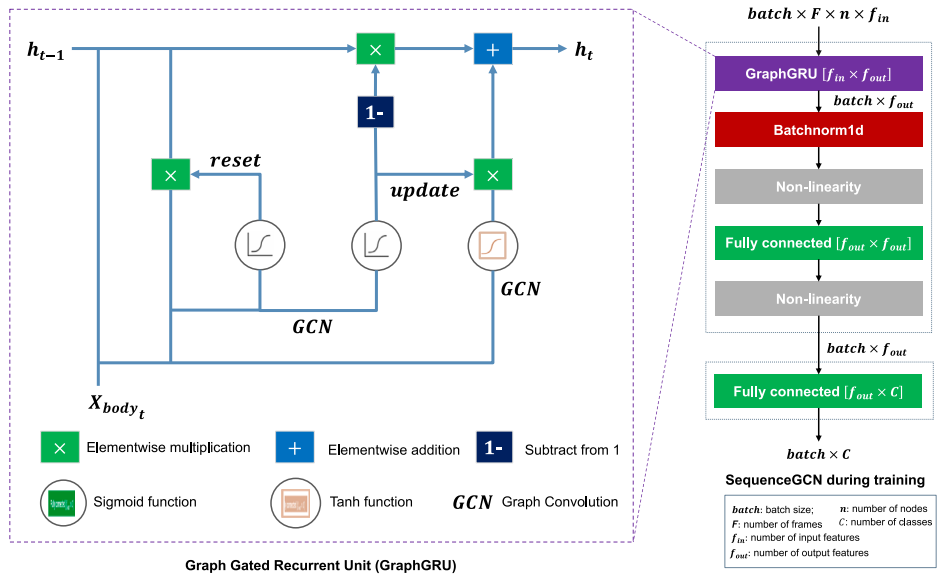

**Fig. 15.** SequenceGCN Architecture. The architecture is based on the Gated Recurrent Unit and takes Gait Graphs as a sequence.

**Table 10**
SequenceGCN results on GITW.

| Method | Accuracy | Class F1 scores | | |
|---|---|---|---|---|
| | | child | adult | senior |
| Gait as a Sequence (SequenceGCN) | 39% | 0.29 | 0.53 | 0.10 |
| Spatiotemporal Gait (PairGCN) | 99% | 0.99 | 0.99 | 0.99 |

### 6.4. Comparison with existing methods

PairGCN is proposed to extract age-associated model-based gait features for age group classification. However, at the time of this writing, existing model-based gait techniques for age group classification make use of handcrafted features and perform classification with conventional classifiers such as Random Forest (Aderinola et al., 2021a; Hema et al., 2019), Support Vector Machines (Chuen et al., 2015; Begg et al., 2005; Hema and Pitta, 2019; Yoo and Kwon, 2017), and K-Nearest Neighbor (Yang and Wang, 2015; Hediyeh et al., 2013).

This section compares features extracted using PairGCN with the performance of handcrafted features, using SVM, KNN, and RF as classifiers. The best-performing settings of the classifiers are chosen for comparison. The results in Table 11 show that compared to manually extracted features, PairGCN features contain richer age-associated features. It is also observed that the accuracy can be further improved with classifiers such as SVM, KNN, or Random Forests.

### 6.5. Discussion

We have proposed and evaluated PairGCN via several experiments on the front and side views of the GITW dataset. We showed that using the side view of gait offers 3% more accuracy because most of the dynamic features of gait can be observed on the side view (Table 7). A comparison of PairGCN to the BaselineGCN also shows that PairGCN's adaptive weight learning reduces the error rate by up to 5%. Moreover, the proposed part aggregation scheme enables faster and better performance for both architectures even without adaptive edge weights.

Body coordinate features are noisy due to false detections and multiple detections of joints during pose estimation. Hence, we performed experiments to observe the effects of smoothing. The experiments suggest that the age-associated features of the elderly are subtle and more sensitive to noise than children, as children can easily be detected by their smaller body sizes and larger head-to-body ratios (Table 9).

Since gait is measured from the walking sequence of humans, it may seem more natural to use sequence-based architectures for gait age prediction. Hence, we examined this experimentally with a SequenceGCN, which was unable to learn age-associated features from gait (Table 10). This suggests that it is best to model gait as a large spatiotemporal graph to capture the inter-frame temporal information, which is crucial in age discrimination. This finding also corroborates this observation that several spatiotemporal gait features are measured from the inter-frame connections of joints.

### 7. Conclusion and future research

In this paper, we have proposed a deep learning approach to extract age-associated features from model-based gait for age group classification, which is the first of its kind. In particular, we proposed a Part-AdaptIve Residual GCN (PairGCN) for automatic extraction of age-associated gait features in video sequences. PairGCN outperformed manual feature selection methods and a Gated Recurrent Unit architecture on a dataset collected in unconstrained environments, achieving up to 99% accuracy in classifying subjects as a child, adult, or senior. These results show the feasibility of age classification in live surveillance. To aid further related research, the dataset collected for gait-based age and gender classification in unconstrained environments is being made publicly available.

There are several limitations in this study. Firstly, the proposed model is sensitive to view change. Though it achieves high accuracy on the side view of gait, the performance degrades on the front view. However, the viewpoint challenge is common to all gait-related tasks. Domain adaptation has been applied in tasks like character recognition, enabling a model to perform comparably well on a dataset slightly different from the dataset it was trained on. Future research can achieve view-invariant gait-based age group classification by using domain adaptation.

In addition, PairGCN validation losses are relatively high (around 20%) due to the noisy nature of coordinates and the variability of age-associated gait features. For example, a child might have a gait like an

**Table 11**
Comparison of manually extracted features with PairGCN features.

| Feature extraction | Works | Final classifier | Acc. | Class F1 scores | | |
|---|---|---|---|---|---|---|
| | | | | child | adult | senior |
| Handcrafted | Chuen et al. (2015), Hema and Pitta (2019) | SVM (linear kernel) | 64% | 0.78 | 0.67 | 0.14 |
| | Begg et al. (2005), Yoo and Kwon (2017) | SVM (RBF kernel) | 74% | 0.85 | 0.72 | 0.62 |
| | Hediyeh et al. (2013) | KNN (k=26) | 73% | 0.82 | 0.73 | 0.54 |
| | Yang and Wang (2015) | KNN (k=5) | 80% | 0.86 | 0.80 | 0.72 |
| | Hema et al. (2019) | RF (100 trees) | 89% | 0.91 | 0.91 | 0.84 |
| | Aderinola et al. (2021a) | RF (150 trees) | 89% | 0.92 | 0.91 | 0.84 |
| Learned | BaselineGCN (You et al., 2020) | RF (150 trees) | 97% | 0.97 | 0.96 | 0.99 |
| | PairGCN (ours) | PairGCN FC | 90% | 0.94 | 0.90 | 0.82 |
| | | SVM (RBF kernel) | 98% | 0.99 | 0.97 | 0.97 |
| | | KNN (k=5) | 96% | 0.99 | 0.95 | 0.93 |
| | | RF (150 trees) | **99%** | **0.99** | **0.99** | **0.99** |

The best values are shown in a **bold** font face. SVM: Support Vector Machine; RBF: Radial Basis Function; KNN: K-Nearest Neighbor; RF: Random Forest.

adult and vice-versa. PairGCN is also prone to overfitting on the gait graphs when 200 epochs are exceeded. Hence, features extracted using PairGCN are best used with classifiers more robust to the irregularities in age-associated features.

To address the limitations above, future research will focus on collecting more data to enable more fine-grained age prediction while exploring domain adaption for view-invariant age group classification.

**CRediT authorship contribution statement**

**Timilehin B. Aderinola:** Formal analysis, Investigation, Writing – original draft. **Tee Connie:** Conceptualization, Data curation, Supervision, Writing – review & editing. **Thian Song Ong:** Data curation, Supervision, Writing – review & editing. **Andrew Beng Jin Teoh:** Resources, Writing – review & editing. **Michael Kah Ong Goh:** Validation, Writing – review & editing.

**Declaration of competing interest**

The authors declare that they have no known competing financial interests or personal relationships that could have appeared to influence the work reported in this paper.

**Data availability**

The data is is available at https://github.com/timiderinola/mmu_gag.

**Acknowledgments**

This work was supported in part by research grants from Fundamental Research Grant Scheme (FRGS) (FRGS/1/2020/ICT02/MMU/02/5) and Multimedia University IR Fund (MMUI/220026).

**References**

Aderinola, T.B., Connie, T., Ong, T.S., Goh, K.O.M., 2021a. Automatic extraction of spatio-temporal gait features for age group classification. In: Garg, L., Sharma, H., Goyal, S.B., Singh, A. (Eds.), Proceedings of International Conference on Innovations in Information and Communication Technologies. Springer, pp. 71–78. http://dx.doi.org/10.1007/978-981-16-0873-5_6.

Aderinola, T.B., Connie, T., Ong, T.S., Yau, W.-C., Teoh, A.B.J., 2021b. Learning age from gait: A survey. IEEE Access 9, 100352–100368. http://dx.doi.org/10.1109/ACCESS.2021.3095477.

Akbari, A., Awais, M., Fatemifar, S., Khalid, S.S., Kittler, J., 2021. A novel ground metric for optimal transport-based chronological age estimation. IEEE Trans. Cybern. 52 (10), 9986–9999.

Begg, R., Palaniswami, M., Owen, B., 2005. Support vector machines for automated gait classification. IEEE Trans. Biomed. Eng. 52 (5), 828–838. http://dx.doi.org/10.1109/TBME.2005.845241.

Berksan, M., 2019. Gender Recognition and Age Estimation Based on Human Gait (Ph.D. thesis). Başkent Üniversitesi Fen Bilimleri Enstitüsü.

Bronstein, M.M., Bruna, J., LeCun, Y., Szlam, A., Vandergheynst, P., 2017. Geometric deep learning: Going beyond euclidean data. IEEE Signal Process. Mag. 34 (4), 18–42. http://dx.doi.org/10.1109/MSP.2017.2693418.

Burt, C., 2020. Biometrics could help gaming industry with cybercrime, age verification as market evolves | biometric update.

Cao, W., Mirjalili, V., Raschka, S., 2020. Rank consistent ordinal regression for neural networks with application to age estimation. Pattern Recognit. Lett. 140, 325–331.

Chuen, B.K.Y., Connie, T., Song, O.T., Goh, M., 2015. A preliminary study of gait-based age estimation techniques. In: 2015 Asia-Pacific Signal and Information Processing Association Annual Summit and Conference (APSIPA). pp. 800–806. http://dx.doi.org/10.1109/APSIPA.2015.7415382.

Cui, Y., Jia, M., Lin, T.-Y., Song, Y., Belongie, S., 2019. Class-balanced loss based on effective number of samples. In: 2019 IEEE/CVF Conference on Computer Vision and Pattern Recognition (CVPR). http://dx.doi.org/10.1109/cvpr.2019.00949.

Dagher, I., Barbara, D., 2021. Facial age estimation using pre-trained CNN and transfer learning. Multimedia Tools Appl. 80, 20369–20380.

Fang, H.-S., Li, J., Tang, H., Xu, C., Zhu, H., Xiu, Y., Li, Y.-L., Lu, C., 2022. Alphapose: Whole-body regional multi-person pose estimation and tracking in real-time. IEEE Trans. Pattern Anal. Mach. Intell..

Han, S., 2020. Age estimation from face images based on deep learning. In: 2020 International Conference on Computing and Data Science. CDS, IEEE, pp. 288–292.

Han, J., Bhanu, B., 2006. Individual recognition using gait energy image. IEEE Trans. Pattern Anal. Mach. Intell. 28 (2), 316–322. http://dx.doi.org/10.1109/TPAMI.2006.38.

Hediyeh, H., Sayed, T., Zaki, M.H., 2013. Use of spatiotemporal parameters of gait for automated classification of pedestrian gender and age. Transp. Res. Rec. 2393 (1), 31–40. http://dx.doi.org/10.3141/2393-04.

Hema, M., Babulu, K., Balaji, N., 2019. Gait based human age classification using random forest classifier. i-manager's J. Pattern Recognit. 6 (2), 1–7.

Hema, M., Pitta, S., 2019. Human age classification based on gait parameters using a gait energy image projection model. In: 2019 3rd International Conference on Trends in Electronics and Informatics (ICOEI). pp. 1163–1168. http://dx.doi.org/10.1109/ICOEI.2019.8862788.

Hochreiter, S., Schmidhuber, J., 1997. Long short-term memory. Neural Comput. 9 (8), 1735–1780.

Islam, T.U., Awasthi, L.K., Garg, U., 2021. Gender and age estimation from gait: A review. In: International Conference on Innovative Computing and Communications: Proceedings of ICICC 2020, Volume 2. Springer, pp. 947–962.

Kitchat, K., Limcharoen, P., Khamsemanan, N., Nattee, C., 2022. Human age estimation from multi-angle gait silhouettes with convolutional neural networks. Thai J. Math. 20 (3), 1227–1238.

Kwasny, D., Hemmerling, D., 2021. Gender and age estimation methods based on speech using deep neural networks. Sensors 21 (14), http://dx.doi.org/10.3390/s21144785.

Lau, L., Chan, K., 2023. Tree structure convolutional neural networks for gait-based gender and age classification. Multimedia Tools Appl. 82 (2), 2145–2164.

Li, N., Zhao, X., Ma, C., 2020. JointsGait:A model-based gait recognition method based on gait graph convolutional networks and joints relationship pyramid mapping. arXiv:2005.08625, [cs, eess].

Liao, R., Cao, C., Garcia, E.B., Yu, S., Huang, Y., 2017. Pose-based temporal-spatial network (PTSN) for gait recognition with carrying and clothing variations. In: Zhou, J., Wang, Y., Sun, Z., Xu, Y., Shen, L., Feng, J., Shan, S., Qiao, Y., Guo, Z., Yu, S. (Eds.), Biometric Recognition. Springer International Publishing, pp. 474–483. http://dx.doi.org/10.1007/978-3-319-69923-3_51.

Liao, H., Yan, Y., Dai, W., Fan, P., 2018. Age estimation of face images based on CNN and divide-and-rule strategy. Math. Probl. Eng..

Liao, R., Yu, S., An, W., Huang, Y., 2020. A model-based gait recognition method with body pose and human prior knowledge. Pattern Recognit. 98, 107069. http://dx.doi.org/10.1016/j.patcog.2019.107069.

Mansouri, N., Issa, M.A., Jemaa, Y.B., 2018. Gait features fusion for efficient automatic age classification. IET Comput. Vis. 12 (1), 69–75. http://dx.doi.org/10.1049/iet-cvi.2017.0055.

Moolla, Y., De Kock, A., Mabuza-Hocquet, G., Ntshangase, C.S., Nelufule, N., Khanyile, P., 2021. Biometric recognition of infants using fingerprint, iris, and ear biometrics. IEEE Access 9, 38269–38286.

Morris, C., Ritzert, M., Fey, M., Hamilton, W.L., Lenssen, J.E., Rattan, G., Grohe, M., 2019. Weisfeiler and leman go neural: Higher-order graph neural networks. In: Proceedings of the AAAI Conference on Artificial Intelligence. Vol. 33, pp. 4602–4609. http://dx.doi.org/10.1609/aaai.v33i01.33014602.

Nabila, M., Mohammed, A.I., Yousra, B.J., 2018. Gait-based human age classification using a silhouette model. IET Biometr. 7 (2), 116–124. http://dx.doi.org/10.1049/iet-bmt.2016.0176.

Phillips, P., Sarkar, S., Robledo, I., Grother, P., Bowyer, K., 2002. Baseline results for the challenge problem of HumanID using gait analysis. In: Proceedings of Fifth IEEE International Conference on Automatic Face Gesture Recognition. pp. 137–142. http://dx.doi.org/10.1109/AFGR.2002.1004145.

Punyani, P., Gupta, R., Kumar, A., 2018. Human age-estimation system based on double-level feature fusion of face and gait images. Int. J. Image Data Fusion 9 (3), 222–236. http://dx.doi.org/10.1080/19479832.2018.1423644.

Rahman, S.T., Arefeen, A., Mridul, S.S., Khan, A.I., Subrina, S., 2020. Human age and gender estimation using facial image processing. In: 2020 IEEE Region 10 Symposium. TENSYMP, IEEE, pp. 1001–1005.

Rani, V., Kumar, M., 2023. Human gait recognition: A systematic review. Multimedia Tools Appl. 1–35.

Rizwan, S.A., Ghadi, Y.Y., Jalal, A., Kim, K., 2022. Automated facial expression recognition and age estimation using deep learning. Comput. Mater. Continua 71 (3).

Russel, N.S., Selvaraj, A., 2021. Gender discrimination, age group classification and carried object recognition from gait energy image using fusion of parallel convolutional neural network. IET Image Process. 15 (1), 239–251. http://dx.doi.org/10.1049/ipr2.12024.

Sarkar, S., Phillips, P., Liu, Z., Vega, I., Grother, P., Bowyer, K., 2005. The humanID gait challenge problem: data sets, performance, and analysis. IEEE Trans. Pattern Anal. Mach. Intell. 27 (2), 162–177. http://dx.doi.org/10.1109/TPAMI.2005.39.

Sharma, N., Sharma, R., Jindal, N., 2022. Face-based age and gender estimation using improved convolutional neural network approach. Wirel. Pers. Commun. 124 (4), 3035–3054.

Sheng, W., Li, X., 2021. Multi-task learning for gait-based identity recognition and emotion recognition using attention enhanced temporal graph convolutional network. Pattern Recognit. 114, 107868.

Shutler, J.D., Grant, M.G., Nixon, M.S., Carter, J.N., 2004. On a large sequence-based human gait database. In: Lotfi, A., Garibaldi, J.M. (Eds.), Applications and Science in Soft Computing. In: Advances in Soft Computing, Springer, pp. 339–346. http://dx.doi.org/10.1007/978-3-540-45240-9_46.

Si, S., Wang, J., Peng, J., Xiao, J., 2022. Towards speaker age estimation with label distribution learning. In: ICASSP 2022 - 2022 IEEE International Conference on Acoustics, Speech and Signal Processing (ICASSP). pp. 4618–4622. http://dx.doi.org/10.1109/ICASSP43922.2022.9746378.

Song, C., Huang, Y., Huang, Y., Jia, N., Wang, L., 2019. GaitNet: An end-to-end network for gait based human identification. Pattern Recognit. 96, 106988. http://dx.doi.org/10.1016/j.patcog.2019.106988.

Song, T., Yang, X., Wang, Y., Lei, Y., Liu, G., 2021. Pedestrian age recognition method based on gait deep learning. J. Phys.: Conf. Ser. 2010 (1), 012031.

Teepe, T., Khan, A., Gilg, J., Herzog, F., Hormann, S., Rigoll, G., 2021. Gaitgraph: Graph convolutional network for skeleton-based gait recognition. In: 2021 IEEE International Conference on Image Processing (ICIP). http://dx.doi.org/10.1109/icip42928.2021.9506717.

Xie, J.-C., Pun, C.-M., 2020. Deep and ordinal ensemble learning for human age estimation from facial images. IEEE Trans. Inf. Forensics Secur. 15, 2361–2374.

Xu, C., Makihara, Y., Liao, R., Niitsuma, H., Li, X., Yagi, Y., Lu, J., 2021a. Real-time gait-based age estimation and gender classification from a single image. In: Proceedings of the IEEE/CVF Winter Conference on Applications of Computer Vision. pp. 3460–3470.

Xu, C., Sakata, A., Makihara, Y., Takemura, N., Muramatsu, D., Yagi, Y., Lu, J., 2021b. Uncertainty-aware gait-based age estimation and its applications. IEEE Trans. Biometr. Behav. Identity Sci. 3 (4), 479–494.

Yaman, D., Irem Eyiokur, F., Kemal Ekenel, H., 2019. Multimodal age and gender classification using ear and profile face images. In: Proceedings of the IEEE/CVF Conference on Computer Vision and Pattern Recognition Workshops.

Yang, C., Wang, W., 2015. A novel age interval identification method based on gait monitoring. In: 2015 4th International Conference on Computer Science and Network Technology (ICCSNT). Vol. 01, pp. 266–269. http://dx.doi.org/10.1109/ICCSNT.2015.7490749.

Yoo, H.W., Kwon, K.Y., 2017. Method for classification of age and gender using gait recognition. Trans. Korean Soc. Mech. Eng. A 41 (11), 1035–1045. http://dx.doi.org/10.3795/KSME-A.2017.41.11.1035.

You, J., Ying, R., Leskovec, J., 2020. Design space for graph neural networks. In: Proceedings of the 34th International Conference on Neural Information Processing Systems. NIPS '20, Curran Associates Inc..

Yu, S., Tan, D., Tan, T., 2006. A framework for evaluating the effect of view angle, clothing and carrying condition on gait recognition. In: 18th International Conference on Pattern Recognition. ICPR'06, Vol. 4, pp. 441–444. http://dx.doi.org/10.1109/ICPR.2006.67.

Zhang, K., Liu, N., Yuan, X., Guo, X., Gao, C., Zhao, Z., Ma, Z., 2018. Fine-grained age estimation in the wild with attention LSTM networks. arXiv:1805.10445, [cs].

Zhang, D., Wang, Y., Bhanu, B., 2010. Age classification base on gait using HMM. In: 2010 20th International Conference on Pattern Recognition. pp. 3834–3837. http://dx.doi.org/10.1109/ICPR.2010.934.

Zhang, S., Wang, Y., Li, A., 2022. Gait energy image-based human attribute recognition using two-branch deep convolutional neural network. IEEE Trans. Biometr. Behav. Identity Sci. 5 (1), 53–63.

Zhao, A., Li, J., Ahmed, M., 2020. SpiderNet: A spiderweb graph neural network for multi-view gait recognition. Knowl.-Based Syst. 206, 106273.

Zhu, H., Zhang, Y., Li, G., Zhang, J., Shan, H., 2020. Ordinal distribution regression for gait-based age estimation. Sci. China Inf. Sci. 63, 1–14.